\documentclass{sig-alternate}

\usepackage[lined,ruled]{algorithm2e}
\usepackage{url}

\begin{document}
%
\conferenceinfo{MM}{'15, October 26-30, 2015, Brisbane, Australia}

\crdata{ISBN 978-1-4503-3459-4/15/10\$15.00 \\DOI: http://dx.doi.org/10.1145/2733373.2806283}  

\title{Sense Beyond Expressions: Cuteness}

\numberofauthors{4} 
%
\author{
%
%
Kang Wang$^*$\\
       \affaddr{Rensselaer Polytechnic Institute}\\
       \email{wangk10@rpi.edu}
\and
Tam V. Nguyen\thanks{indicates the co-first authorship}\\
       \affaddr{Singapore Polytechnic}\\
       \email{nguyen\_van\_tam@sp.edu.sg}
\and
Jiashi Feng\\
       \affaddr{University of California, Berkeley}\\
		\email{jshfeng@gmail.com}       
\and  
Jose Sepulveda\\
       \affaddr{Singapore Polytechnic}\\
       \email{sepulveda\_jose@sp.edu.sg}
}

\maketitle
\begin{abstract}
With the development of Internet culture, \emph{cute} has become a popular concept. Many people are curious about what factors making a person look cute. However, there is rare research to answer this interesting question. In this work, we construct a dataset of personal images with comprehensively annotated cuteness scores and facial attributes to investigate this high-level concept in depth. Based on this dataset, through an automatic attributes mining process, we find several critical attributes determining the cuteness of a person. We also develop a novel Continuous Latent Support Vector Machine (C-LSVM) method to predict the cuteness score of one person given only his image. Extensive evaluations validate the effectiveness of the proposed method for cuteness prediction.
\end{abstract}




\section{Introduction}
Cuteness describes a type of attractiveness commonly associated with youth and appearance, which activates in others the motivation to care~\cite{lorenz1971part}. Recent studies suggest that cute images stimulate the pleasure centers of the brain which is closely related with the positive emotion of human~\cite{glocker2009baby}. This explains why everybody prefers \textit{cute} persons or stuff in social network, shopping, browsing images/videos on the web and so on. For example, some survey shows that women's fashions opt for the cute even over the sensible or glamorous\footnote{\small \url{http://www.nytimes.com/2006/01/03/science/03cute.html?pagewanted=all}}. This makes cuteness be a quite important factor to consider in product design, advertisement and so on.

\begin{figure}[t]
\includegraphics[scale=0.35]{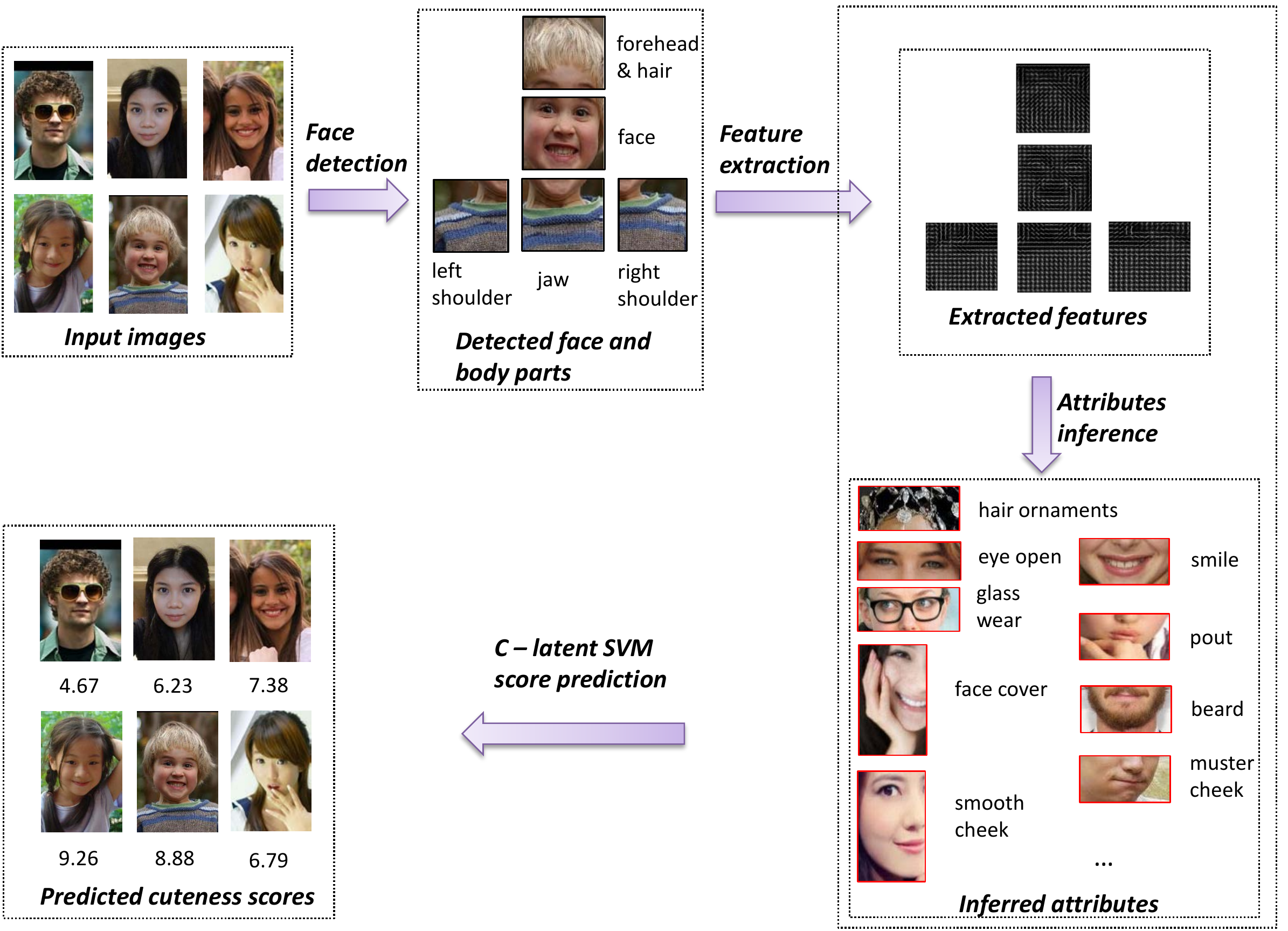}
\vspace{-7mm}
\caption{The framework of cuteness prediction. For each input image, we first detect the bounding box of the human faces and neighboring body parts, and then extract the appearance features accordingly. Based on the appearance features, we infer the underlying middle-level attributes and the cuteness score simultaneously.}
\label{fig:framework}
\end{figure}

Cuteness has received the attention of psychologists and neuroscientists for several years. For example, Kim \emph{et al.}~\cite{kim2007neural} conducted studies on why humans think certain animals are cute using functional magnetic resonance imaging (fMRI) to measure changes in human brain activity. Some evidence from this work suggests the brain activity is greater when the stimulus has juvenile characteristics~--~a button nose, big eyes, a large wobbly head, fat cheeks, \emph{etc}.

Though it has been investigated by psychologists and neuroscientists, cuteness has not roused the attention of computer scientists yet. In computer vision and multimedia files, there are plenty of works focusing on recognizing ordinary expressions, \textit{i.e.}, happiness and sadness. However, cuteness, which is beyond these traditional expressions and has higher-level semantics, is far more difficult to recognize. In this work, we explore the secrets of cuteness through the application of machine learning techniques. This is the first research attempt of computational analysis on what factors determine the cuteness, a high-level concept. We construct a model that learns from human images and their respective cuteness ratings to produce human-like evaluation of cuteness. Our work is based on the underlying theory that there are objective regularities in cuteness to be analyzed and learned. And in this work, we also provide a general framework for investigating and analyzing other high-level expressions, such as ``funny'' and ``scary''.

To investigate the factors determining the cuteness of a person, we construct a large dataset of human images with comprehensively annotated attributes and cuteness scores. In this work, the attributes are defined empirically and used as the middle-level descriptors of certain characteristic of the persons. For example, some of the attributes describe the facial appearance such as \emph{skin smoothness} and \emph{age}. And others describe the pose or expression of the human, such as \emph{smile} and \emph{face cover}. Based on this dataset, we propose a novel model to automatically learn which features and attributes determine the cuteness of persons. And we train the predictors on these features and attributes for predicting the cuteness score of a new person image. In previous works, latent SVM is a widely adopted method for the attributes mining and prediction~\cite{liu2012hi,wang2010discriminative}. However, for the cuteness prediction problem, the annotations of the samples (cuteness scores) are continuous values. Thus, the traditional latent SVM, which can only handle discrete annotations, cannot be applied here. In this work, we propose a novel Continuous Latent SVM (C-LSVM) method, which can handle the continuous labels of the samples, to solve this issue. And we show that C-LSVM is a more general method than standard Latent SVM and has a great potential for solving many other problems involving predicting continuous variables.


Studying what makes person looks cute and how to predict cuteness from person images alone may have many useful applications. These applications include choosing a collection of cute clips from a video to generate the attractive video summarization, organizing photos in an album according to the cuteness, automatically retrieve the cute images in a large image set. Cuteness prediction and generation also benefit greatly the advertisement and production design.

\section{Dataset Construction}
\label{sec:dataset}

Since none datasets exist for the cuteness research, in order to investigate this problem well, we collect a new dataset of personal images by ourselves. The images are crawled from the web and $4,800$ images are collected in total. Most of these images contain the frontal face of the persons.

\begin{figure}[t]
\centering
\includegraphics[width=\linewidth]{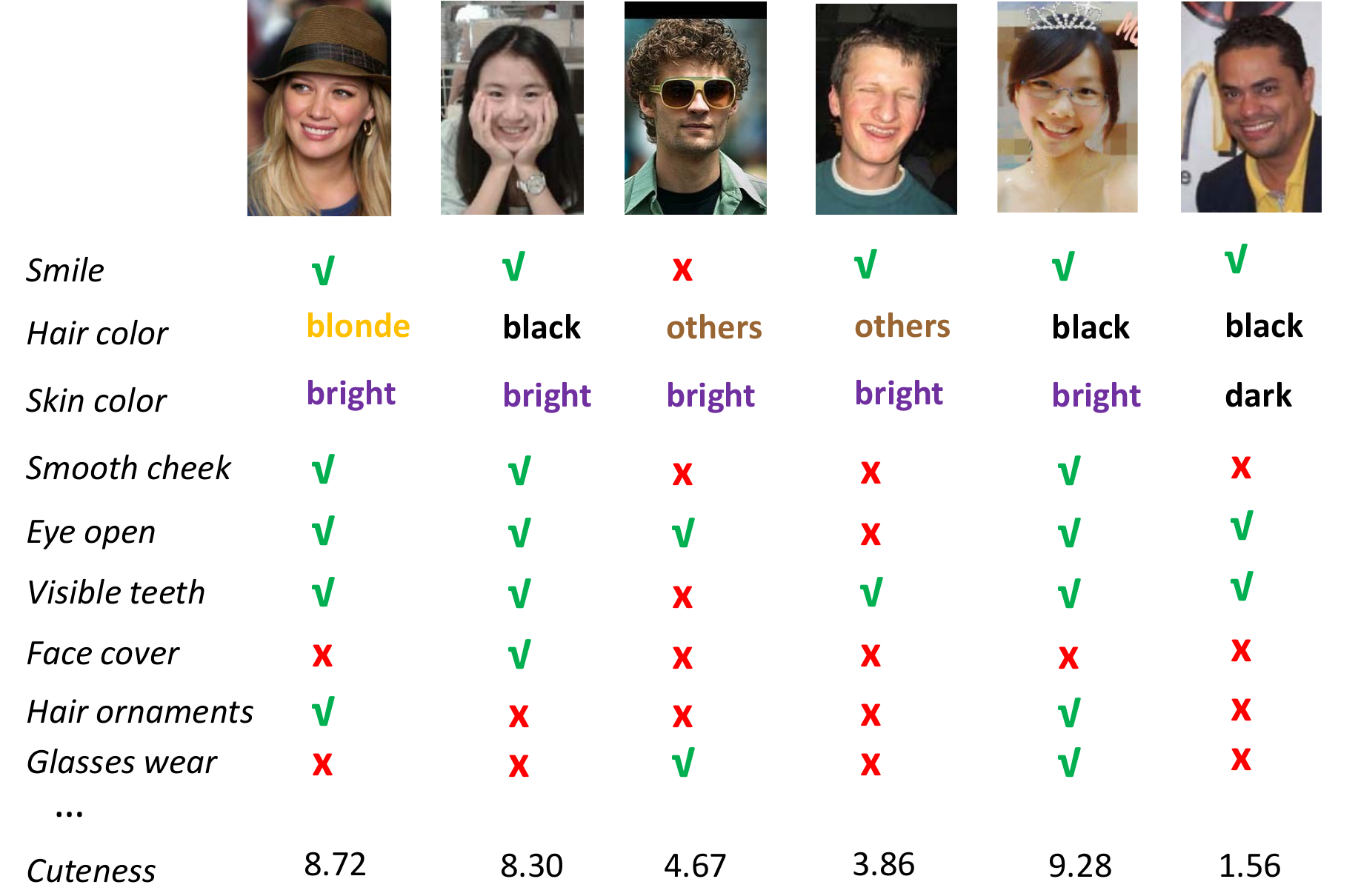}
\vspace{-7mm}
\caption{Examples of the collected photos with annotated attributes and cuteness scores.}
\label{fig:examples}
\end{figure}

We invite $40$ subjects to participate the annotation of the images' scores and attributes.
To relieve the burden of the cuteness score annotation, in this work, we adopt a $k$-wise comparison to estimate the rank of the images and then automatically infer their absolute scores~\cite{vansense,TamNguyen}. In each round of annotation, the subjects are required to rank $k$ images in a descend order of their scores. After they finish all the annotations, we estimate the absolute score of the cuteness based a rank SVM method, which maintains the rank of the photos annotated in each $k$-wise annotation. For the details please refer to~\cite{vansense,TamNguyen}.
The cuteness degree of a person heavily depends on his appearance and pose. For example, a young and pretty girl pouting her mouth will look quite cute. In this work, we define following $19$ attributes to comprehensively describe the appearance and pose of a person in a middle semantic level. The adopted attributes include Gender (Male, Female), Age (Young, Teen, Middle, Old), Eye (Open, Close), Mouth Variation, Mouth (Open, Close), Teeth (Visible, Invisible), Smile, Wearing Glasses, Beard, Skin Color (Bright, Dark), Hair Color (Black, Blonde, Other), Hair Ornaments, Face Cover, Skin Smoothness. For the attribute of mouth variation, we consider the poses such as pout, muster cheek having the value of $1$. For the attribute of face cover, it includes finger touches lips, hand holds jaw and so on. Some examples of the above attributes are shown in Figure~\ref{fig:framework} and Figure~\ref{fig:examples}. Most of the attributes admit two values and are represented by a binary variable. 


\section{C-LSVM for Cuteness Prediction}
\label{sec:method}
\subsection{Features}
We first run Viola-Jones face detector~\cite{viola2004robust} on the collected images to obtain bounding boxes for the human faces. Then we extract the appearance features from the face bounding box and the four spatially neighboring bounding boxes with the same size as shown in Figure~\ref{fig:framework}. Here all of the bounding boxes are resized to $128\times 128$ pixels. We extract Gabor, LBP~\cite{ahonen2006face}, HOG~\cite{dalal2005histograms} features within the bounding boxes to describe the pose and facial texture of persons. More specifically, for the Gabor feature, we extract the filter responses in $5$ scales and $8$ orientations and form a $128\times 128\times 5\times 8$ feature vector for each bounding box. Then its feature dimension is reduced to $200$ dimensions via PCA. Similarly, we extract HOG feature and reduce its dimensionality from $8,000$ to $200$ and LBP feature from $256$ to $40$. We experimentally find that reducing the feature dimensionality can enhance both the prediction performance and efficiency.

\subsection{Prediction Model}
We propose a continuous latent support vector machine (C-LSVM) for modeling the relationship between the raw feature with the attributes and cuteness score. In particular, we use three linear models to describe the relationship between raw feature $\mathbf{x}$ and attribute $\mathbf{a}$, raw feature $\mathbf{x}$ and cuteness score $y$, attributes $\mathbf{a}$ and cuteness score $y$ respectively. For an image $I_i$, we extract the aforementioned features $\mathbf{x}_i$. The relationship of the attributes and cuteness score with the raw feature are modeled as follows:
\begin{eqnarray*}
y_i &=& \mathbf{w}_{x,y}^T \mathbf{x}_i + b_{x,y}, \\
\mathbf{a}_i &=& W_{x,a}^T \mathbf{x}_i + b_{x,a}, \\
y_i &=& \mathbf{w}_{a,y}^T \mathbf{a}_i + b_{a,y}.
\end{eqnarray*}
Here $\mathbf{w}_{x,y}, b_{x,y}, W_{x,a}, b_{x,a}, \mathbf{w}_{a,y}, b_{a,y}$ are the parameters of the linear prediction models, which will be determined in the model learning process. In the proposed C-LSVM method, the cuteness score of an image is inferred by maximizing the following fitness function,
\begin{eqnarray}
\label{eqn:loss_function}
\mathcal{F} &=& - \beta_1 (\mathbf{w}_{x,y}^T\mathbf{x}_i + b_{x,y}-y)^2 - \beta_2 (\mathbf{w}_{a,y}^T\mathbf{a} + b_{a,y}-y)^2 \nonumber \\
&& - \|\Lambda(W_{x,a}^T\mathbf{x}_i + \mathbf{b}_{x,a}-\mathbf{a})\|^2 + \mathbf{a}^T(P\otimes M)\mathbf{a}.
\end{eqnarray}
The fitness function is a linear combination of the following three types of fitness, \emph{i.e.}, score prediction from raw feature, score prediction from attributes, the attributes prediction from raw feature. And the last term accounts for the attributes correlation, which encourages the correlated attributes to be predicted simultaneously and can help improve the attributes prediction accuracy. The matrix $M$, constructed based on the statistics from the training data, is of size $n\times n$ and accounts for the attributes co-occurrence.

Besides the three linear models, there are also parameters trading-off the cost terms including $\beta_1,\beta_2,\Lambda,P$. $\Lambda$ is a diagonal matrix with the size of $n\times n$. The $i$th element in the diagonal weights the prediction cost for the $i$th attribute, which in fact reflects the importance of the $i$th attribute in the cuteness score prediction.  And $P$ is a matrix with the same size as $M$, which weights the co-occurrence of each pair of attributes. All of these parameters, including $\beta_1$, $\beta_2$, $\Lambda$ and $P$ need to be determined in the learning process. Here we also employ a latent max-margin framework for the parameter learning and the details are provided in the following subsections.

\subsection{Model Learning}
In the learning process, we construct the following three prediction functions: the first one predicts the cuteness score from the raw feature, $\mathbf{w}_{x,y}: \mathbf{x}\rightarrow y$; the second one predicts the attributes from the raw feature, $W_{x,a}: \mathbf{x} \rightarrow \mathbf{a}$; the third one predicts the cuteness score from the attributes, $\mathbf{w}_{a,y}: \mathbf{a}\rightarrow y$. We adopt a max-margin regression scheme to learn the above three prediction functions individually.
\begin{eqnarray}
\label{eqn:model_learning}
&& \min \frac{1}{2} \|\mathbf{w}_{x,y}\|^2 + C_{x,y} \sum_{i=1}^m (\xi_i + \xi_i^*) \nonumber \\
 \text{s.t.  } &&  y_i - \langle \mathbf{w}_{x,y}, x_i \rangle - b_{x,y} \leq \varepsilon + \xi_i \\
 &&  \langle \mathbf{w}_{x,y}, x_i \rangle + b_{x,y} - y_i \leq \varepsilon + \xi_i^* \nonumber  \\
 && \xi_i, \xi_i^* \geq 0 \nonumber
\end{eqnarray}
After optimizing the above objective function via off-the-shelf solvers, we can obtain the parameters $\mathbf{w}_{x,y},b_{x,y}$ of the prediction function $\phi_{x,y}$. The other two prediction functions, $W_{x,a}, b_{x,a}, \mathbf{w}_{a,y}, b_{a,y}$, can be obtained in a similar way.

For the parameters optimization, including $\beta_1,\beta_2, \lambda, P$, we propose the following C-LSVM method. The objective function is defined as:
\begin{eqnarray*}
&& \min \frac{\gamma}{2}\|\mathbf{z}\|^2 + \sum_{i=1}^m \zeta_i \\
&& \text{s.t. } \max_{a \in \mathcal{S}} \mathbf{z}^T \phi(\mathbf{x}_i,\mathbf{a}, y_i) \geq  \max_{\substack{a \in \mathcal{S}\\ \bar{y} \in \mathcal{Y}_i} } \mathbf{z}^T \phi(\mathbf{x}_i,\mathbf{a}, \bar{y}) + \Delta(y_i,\bar{y}) - \zeta_i,
\end{eqnarray*}
where the set $\mathcal{Y}_i$ is defined as $\mathcal{Y}_i = \{y| y < y_i - \epsilon\} \cup \{ y| y > y_i + \epsilon\}$ containing all of the incorrect cuteness scores. $\epsilon$ is a parameter to control the tolerance of the prediction error and is set as $0.5$ throughout the experiments. And the set $\mathcal{A}$ is defined as $\mathcal{A}=\{\mathbf{a}|0\leq a_i \leq 1,\forall i=1,\ldots,n\}$. Here the parameter $\mathbf{z}$ is the concatenation of $\beta_1,\beta_2,\lambda,P$. The potential function $\phi(\mathbf{x}_i, \mathbf{a}, y)$ is defined as:
\begin{equation*}
    \phi(\mathbf{x}_i, \mathbf{a}, y) = [\phi_{x,y}(\mathbf{x}_i,y); \phi_{a,y}(\mathbf{a},y); \phi_{x,a}(\mathbf{x}_i,\mathbf{a}); \phi_{a,a}(\mathbf{a},\mathbf{a})].
\end{equation*}
In particular, the contained four potential functions are defined as follows:
\begin{eqnarray*}
\phi_{x,y}(\mathbf{x}_i,y) &=& -(\mathbf{w}_{x,y}^T\mathbf{x}_i + b_{x,y}-y)^2,\\
\phi_{a,y}(\mathbf{a},y) &=& -(\mathbf{w}_{a,y}^T\mathbf{a} + b_{a,y}-y)^2,\\
\phi_{x,a}(\mathbf{x}_i,\mathbf{a}) &=& -(W_{x,a}^T\mathbf{x}_i + \mathbf{b}_{x,a}-\mathbf{a})\otimes(W_{x,a}^T\mathbf{x}_i + \mathbf{b}_{x,a}-\mathbf{a}), \\
\phi_{a,a}(\mathbf{a},\mathbf{a}) &=& \mathbf{a}^TP\otimes M\mathbf{a}.
\end{eqnarray*}
The above optimization problem is equivalent to minimizing the following loss function:
\begin{equation}
\label{eqn:parameter_learning}
    \mathcal{L}(\mathbf{z}) = \frac{\gamma}{2}\|\mathbf{z}\|^2 + R(\mathbf{z}),
\end{equation}
where
\begin{equation}
\label{eqn:risk_function}
R(\mathbf{z}) =  \max_{\substack{a \in \mathcal{S}\\ \bar{y} \in \mathcal{Y}_i} } \mathbf{z}^T \phi(\mathbf{x}_i,\mathbf{a}, \bar{y}) + \Delta(y_i,\bar{y}) - \max_{a \in \mathcal{S}} \mathbf{z}^T \phi(\mathbf{x}_i,\mathbf{a}, y_i)
\end{equation}
The subgradient for the above function can be calculated as follows,
\begin{eqnarray*}
\partial L(\mathbf{z}) &=& \mathbf{z} + [\phi_{x,y}(\mathbf{x},y); \phi_{a,y}(\mathbf{a}^*,y); \phi_{x,a}(\mathbf{x},\mathbf{a}^*); \mathbf{m} \otimes \tilde{\mathbf{a}}] \\
&&- [\phi_{x,y}(\mathbf{x},y); \phi_{a,y}(\mathbf{a}^\star,y); \phi_{x,a}(\mathbf{x},\mathbf{a}^\star); \mathbf{m} \otimes \bar{\mathbf{a}}].
\end{eqnarray*}
Here $\mathbf{a}^*$ and $\mathbf{a}^\star$ are obtained from solving the first and second maximizing problems in~\eqref{eqn:risk_function} respectively. Note that the optimization problems are standard quadratic programming and can be solved efficiently.
Here $\mathbf{m}$ is the long vector formed by stacking the column vectors of matrix $M$. $\tilde{\mathbf{a}}$ and $\bar{\mathbf{a}}$ are formed by stacking the column of matrices $\mathbf{a}^*{\mathbf{a}^*}^T$ and $\mathbf{a}^\star{\mathbf{a}^\star}^T$. In the optimization, we alternatively solve the problems~\eqref{eqn:model_learning} and~\eqref{eqn:parameter_learning}. In particular, the problem~\eqref{eqn:model_learning} can be solved via standard SVR solver and we can obtain the individual prediction model. Then the problem~\eqref{eqn:parameter_learning} is solved via standard quadratic programming and the estimation of the underlying attributes are updated. This procedure is repeated until convergence.


\subsection{Inference}
After learning the prediction functions and the model parameters, for a new image, its cuteness score can be inferred as follows,
\begin{eqnarray*}
\{ \mathbf{a},y \} = \arg \max_{\mathbf{a},y} \mathbf{z}^T\phi(\mathbf{x},\mathbf{a},y).
\end{eqnarray*}
In solving the above optimization problem, we first infer the attributes confidence vector $\mathbf{a}$. Then we binarize the elements in $\mathbf{a}$ via a fixed threshold $0.5$\footnote{\small It is widely used in applying regression for classification.}. After determining the attributes, we then infer the cuteness score $y$ via solving a standard quadratic programming problem. This optimization procedure is inspired by the fact that a sparse attribute vector $\mathbf{a}$ generally produces better prediction result of the cuteness score. The rationale lies on that in the learning process, the attributes annotation associated with each sample are binary values. Thus the learned model prefers such sparse input attribute vectors.

\section{Experiments}
\label{sec:experiments}
In this section, we evaluate the proposed C-LSVM cuteness score prediction on the constructed dataset. In the experiments, $3,000$ images from the dataset are used for training.
We compare the C-LSVM method with the following three methods: the first one is based on constructing a feed forward neural network, which has been successfully applied in the beauty prediction~\cite{vansense}. The second one is to directly predict the cuteness score from the raw feature, where the prediction is based on a support vector regression model; and the third one is to predict the attributes at first and then predict the score from the estimated attributes based on two individual SVR models.

\begin{figure}[t]
\centering
\includegraphics[width=0.98\linewidth]{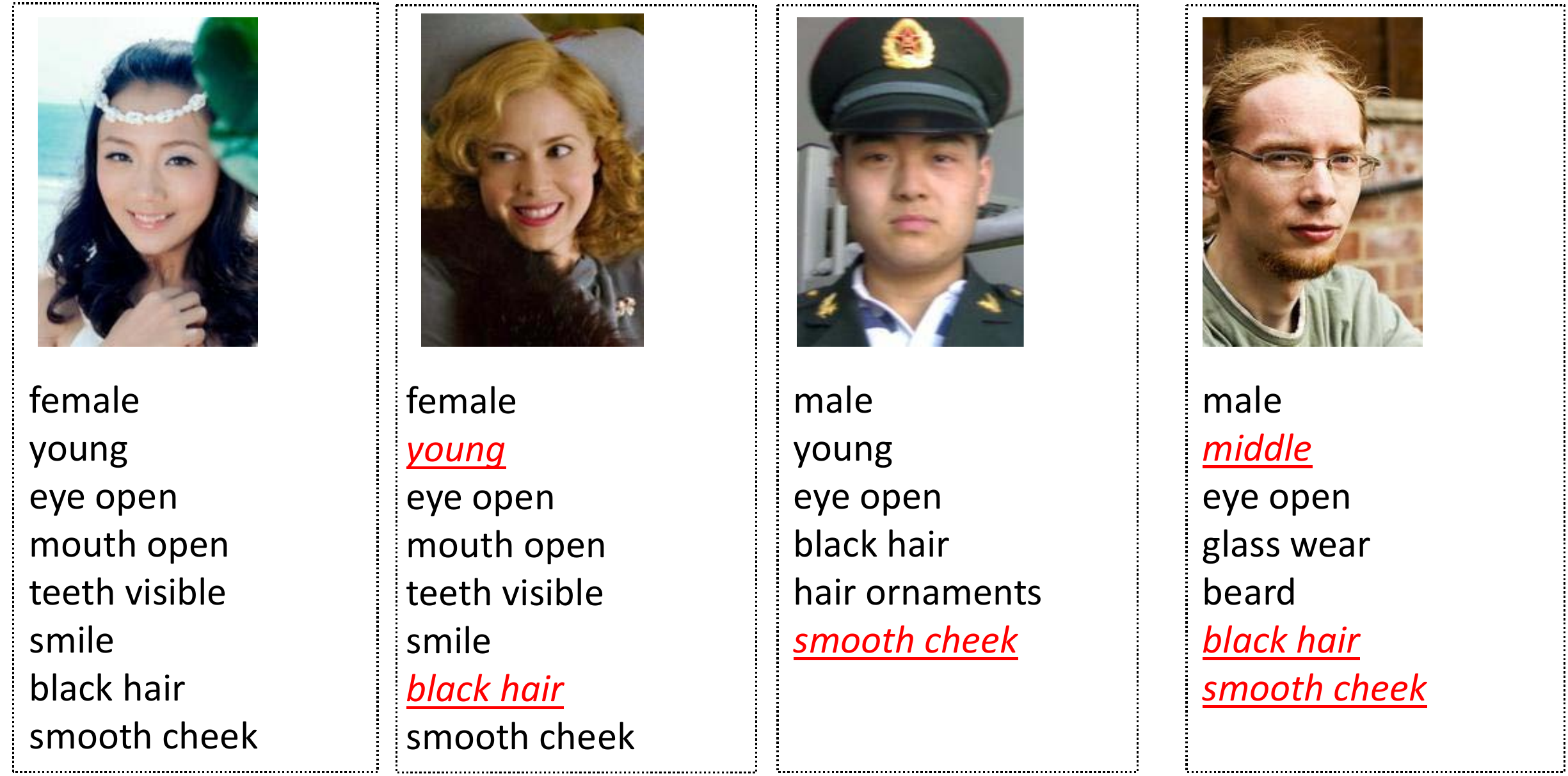}
\vspace{-4mm}
\caption{Examples of the inferred attributes. Most of the attributes can be inferred correctly. And the incorrect inferred attributes are highlighted by red color.}
\label{fig:accAttribute}

\end{figure}

\begin{figure}[t]
\begin{center}
\includegraphics[width=\linewidth]{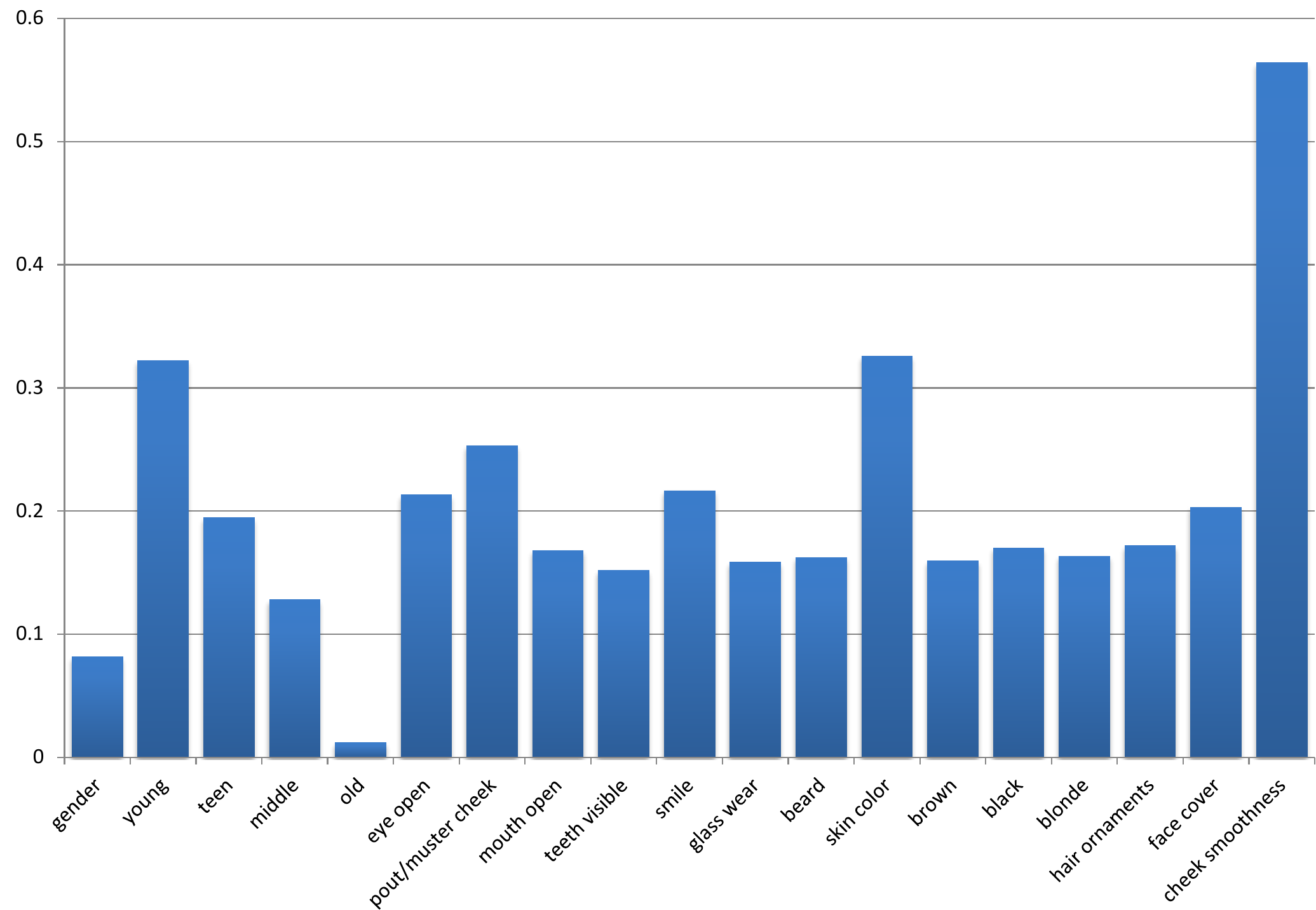}
\end{center}
\vspace{-6mm}
\caption{The mined weights of the attributes for cuteness prediction. The horizontal axis displays the attributes and the vertical axis shows the learned weights.}
\label{fig:attributes_weight}

\end{figure}

The accuracy of the cuteness score prediction is measured by the mean absolute error (MAE) in the evaluation. Note that the groundtruth score is ranged from $0$ to $10$. The evaluation results are presented in the Table 1. From the results, we can observe that the neural network method performs worst. Introducing the attributes will improve the results over only using raw feature by $0.14$. And our proposed C-LSVM can further reduce the MAE by $0.08$ and achieves the best result. Note that the MAE for the last three methods are quite small and such improvement is in fact significant. We also present some attributes inference results for the test samples in Figure~\ref{fig:accAttribute}. We can see that most of the attributes can be correctly inferred. While for the \emph{cheek smoothness}, the accuracy is relatively low. The reason may be that the low-level feature we adopted are describing the whole face region, instead of only describing the cheek region. Thus some noise may contaminate the prediction for the smoothness. To more intuitively show how the defined attributes determine the cuteness of one person, we visualize the learned inference model $\mathbf{w}_{a,y}$ in Figure~\ref{fig:attributes_weight}. We observe that the attribute \emph{cheek smoothness} is most important for the cuteness of one person. And the attributes \emph{age (young)} and \emph{skin color} are also important. Meanwhile, the \emph{age (old)} and \emph{gender} attributes are least important for the cuteness.


\begin{table}
\small
\center
\caption{MAE of the cuteness score prediction.}
\label{Table:MonSeg2}
\begin{tabular}{c c c c c }
\hline
\hline
Method &NN &F-S &F-A-S &C-LSVM  \\
\hline
MAE &$1.92$ &$1.49$ &$1.35$ &$\mathbf{1.27}$  \\
\hline
\end{tabular}
\vspace{-4mm}
\end{table}

%
%

\section{Conclusions}
\label{sec:conclusions}
In this work, we present the first research attempt of computational analysis on the cuteness, which is beyond the ordinary expressions such as happiness or sadness. We construct a large dataset of persons' images with well annotated cuteness scores and attributes. We propose the novel C-LSVM method which automatically mines the important features and attributes determining the cuteness of a person. Extensive evaluations show that our method can better capture the relationship between the raw feature, attributes and the cuteness score, over the traditional linear predictors.

\section{Acknowledgments}
This work was supported by Singapore Ministry of Education
under research Grant MOE2014-TIF-1-G-007.

%
\scriptsize 
\bibliographystyle{abbrv}
\bibliography{facialExpression}  
%
%
\end{document}